\documentclass[pdflatex,sn-nature]{sn-jnl}

\usepackage{graphicx}%
\usepackage{multirow}%
\usepackage{amsmath,amssymb,amsfonts}%
\usepackage{amsthm}%
\usepackage{mathrsfs}%
\usepackage[title]{appendix}%
\usepackage{xcolor}%
\usepackage{textcomp}%
\usepackage{manyfoot}%
\usepackage{booktabs}%
\usepackage{algorithm}%
\usepackage{algorithmicx}%
\usepackage{algpseudocode}%
\usepackage{listings}%
\usepackage{tabularx}
\usepackage{lineno}

\theoremstyle{thmstyleone}%
\theoremstyle{thmstyletwo}%
\theoremstyle{thmstylethree}%

\usepackage{amsmath,amssymb,amsthm,mathtools,bm,bbm}

\usepackage{tikz,pgfplots,pgfplotstable}
\pgfplotsset{compat=newest}
\usetikzlibrary{arrows.meta,positioning,calc,fit,decorations.pathreplacing,patterns}

\usepackage{quantikz}
\usepackage{caption,subcaption}

\usepackage{booktabs}
\usepackage{array} 

\usepackage{amsthm} 

\begin{document}
\title[Article Title]{MediQ-GAN: Quantum-Inspired GAN for High Resolution Medical Image Generation}


\author[1]{\fnm{Qingyue} \sur{Jiao}}\email{qjiao@nd.edu}
\equalcont{These authors contributed equally to this work.}

\author[4]{\fnm{Yongcan} \sur{Tang}}\email{yt2859@columbia.edu}
\equalcont{These authors contributed equally to this work.}

\author[2]{\fnm{Jun} \sur{Zhuang}}\email{junzhuang@boisestate.edu}

\author[3]{\fnm{Jason} \sur{Cong}}\email{cong@cs.ucla.edu}

\author*[1]{\fnm{Yiyu} \sur{Shi}}\email{yshi4@nd.edu}

\affil*[1]{\orgdiv{Department of Computer Science and Engineering}, \orgname{University of Notre Dame}, \orgaddress{\street{Holy Cross Dr.}, \city{Notre Dame}, \postcode{46556}, \state{Indiana}, \country{USA}}}

\affil[2]{\orgdiv{Department of Computer Science}, \orgname{Boise State University}, \orgaddress{\street{1910 W University Dr.}, \city{Boise}, \postcode{83725}, \state{Idaho}, \country{USA}}}

\affil[3]{\orgdiv{Computer Science Department}, \orgname{University of California, Los Angeles}, \orgaddress{\street{405 Hilgard Avenue}, \city{Los Angeles}, \postcode{90095}, \state{California}, \country{USA}}}

\affil[4]{\orgname{Independent Researcher}}



\abstract{Machine learning-assisted diagnosis shows promise, yet medical imaging datasets are often scarce, imbalanced, and constrained by privacy, making data augmentation essential. Classical generative models typically demand extensive computational and sample resources. Quantum computing offers a promising alternative, but existing quantum-based image generation methods remain limited in scale and often face barren plateaus. We present MediQ-GAN, a quantum-inspired GAN with prototype-guided skip connections and a dual-stream generator that fuses classical and quantum-inspired branches. Its variational quantum circuits inherently preserve full-rank mappings, avoid rank collapse, and are theory-guided to balance expressivity with trainability. Beyond generation quality, we provide the first latent-geometry and rank-based analysis of quantum-inspired GANs, offering theoretical insight into their performance. Across three medical imaging datasets, MediQ-GAN outperforms state-of-the-art GANs and diffusion models. While validated on IBM hardware for robustness, our contribution is hardware-agnostic, offering a scalable and data-efficient framework for medical image generation and augmentation.}




\maketitle
Medical imaging datasets are often much smaller than large natural image collections. The problem is even more pronounced for rare diseases, where collecting labeled data is costly, time-consuming, and often constrained by patient availability or ethical considerations. In the absence of sufficient training samples, it is difficult for machine learning-based diagnostic tools to achieve stable and trustworthy performance. To address this scarcity, data augmentation is frequently employed to improve the effectiveness of machine learning-based tasks. Augmentation strategies range from classical image transformations (e.g., flipping, rotation, and scaling) to advanced deep learning–based methods such as generative models. Classical generative approaches, including diffusion-based models \cite{rombach2022ldm} and Generative Adversarial Networks (GANs) \cite{karras2021stylegan3}, have achieved remarkable success across diverse applications. However, their performance typically relies on large volumes of training data and substantial computational resources, often accessible only through public cloud infrastructure. This dependency poses a fundamental limitation: when applied to medical datasets that are inherently limited, sensitive, or imbalanced, classical generative models struggle to produce high-quality generated images.

Quantum computing has emerged as a new computational paradigm with the potential to achieve exponential speedups for problems that are intractable on classical hardware. Beyond its advantages in physics and optimization, quantum computing has recently gained attention for its ability to enhance machine learning by offering large feature spaces and higher expressivity~\cite{Liu2024EfficientQuantum, PhysRevLett.126.190505}. This has given rise to quantum machine learning (QML), which explores how quantum-inspired principles and quantum algorithms can improve learning efficiency and generalization, especially in data-scarce regimes~\cite{Caro2022GeneralizationQML}. Within QML, quantum generative models have been proposed for both quantum data~\cite{Zoufal2019-vj} and classical data~\cite{Silver_2023_ICCV, 9605352}, demonstrating promise for simulation, sampling, and image generation.

Despite these advances, two major challenges remain. First, classical generative adversarial networks (GANs) have demonstrated state-of-the-art performance in image generation~\cite{karras2021stylegan3}, but their training demands large datasets and computational resources~\cite{rombach2022ldm}. Second, existing hybrid classical-quantum generative models are often constrained by the noise and limited scale of current Noisy Intermediate-Scale Quantum (NISQ) devices, restricting their ability to generate high-resolution, high-fidelity images. To address these challenges, we introduce MediQ-GAN, a quantum-inspired GAN framework that combines the scalability of classical convolutional architectures with the expressive power of quantum-inspired modules designed to preserve full-rank mappings and mitigate generator rank collapse.

Specifically, we present a novel skip-connection prototype-based quantum-inspired Generative Adversarial Network that offers superior image generation quality and downstream task performance boost with sample and resource efficiency. Our work makes three contributions to quantum-inspired generative modeling for medical imaging. First, we introduce a dual-stream, quantum-inspired generator that integrates quantum and classical feature pathways through skip connections to generate high-fidelity medical images. Second, we use the generated images for data augmentation, boosting downstream classification accuracy on class-imbalanced datasets. This approach not only supports the creation of extensive medical datasets but also promotes balanced representation across variables such as underlying pathologies and demographic groups. Third, we provide an explanation for the performance gains of MediQ-GAN, which has been largely absent in prior work. We analyze the latent space geometry and show that MediQ-GAN utilizes a larger fraction of the latent space, leading to improved generation quality. We further prove that our quantum circuit ansatz avoids barren plateaus, remains well-conditioned, and achieves an optimal balance between trainability and expressivity.

The proposed MediQ-GAN framework is evaluated on three medical imaging benchmarks: ISIC 2019, a large dermatoscopic dataset for skin lesions~\cite{Tschandl2018TheHD, codella2018skinlesionanalysismelanoma, combalia2019bcn20000dermoscopiclesionswild}, an ophthalmic dataset of color fundus images~\cite{10.1007/978-3-030-71058-3_11}, and RetinaMNIST from MedMNIST~\cite{medmnistv2, Liu2022DeepDRiD}, derived from fundus images originally designed for ordinal regression of diabetic retinopathy severity. Performance is compared against classical generative baselines, including DCGAN~\cite{radford2016unsupervisedrepresentationlearningdeep}, WGAN-GP~\cite{10.5555/3295222.3295327}, Structure-Preserving GANs~\cite{DBLP:conf/icml/BirrellKRZ22}, StyleGAN2 with Adaptive Discriminator Augmentation (ADA)~\cite{NEURIPS2020_8d30aa96}, FastGAN~\cite{liu2021towards}, fine-tuned Diffusion Transformers (DiT)~\cite{10377858}, and LoRA-tuned Stable Diffusion 3~\cite{10.5555/3692070.3692573}, as well as the state-of-the-art hybrid classical-quantum model MosaiQ~\cite{Silver_2023_ICCV}.  Both qualitative inspection and quantitative metrics confirm that MediQ-GAN generates images of superior fidelity and diversity under limited training data. Furthermore, augmenting classifiers with images generated by our model leads to consistent improvements in downstream classification accuracy and AUC score, demonstrating its practical value for rare-disease medical datasets. 

\section*{Results}\label{result}
\begin{figure}[t]
  \centering
  \begin{subfigure}{\textwidth}
    \centering
    \includegraphics[width=0.8\linewidth]{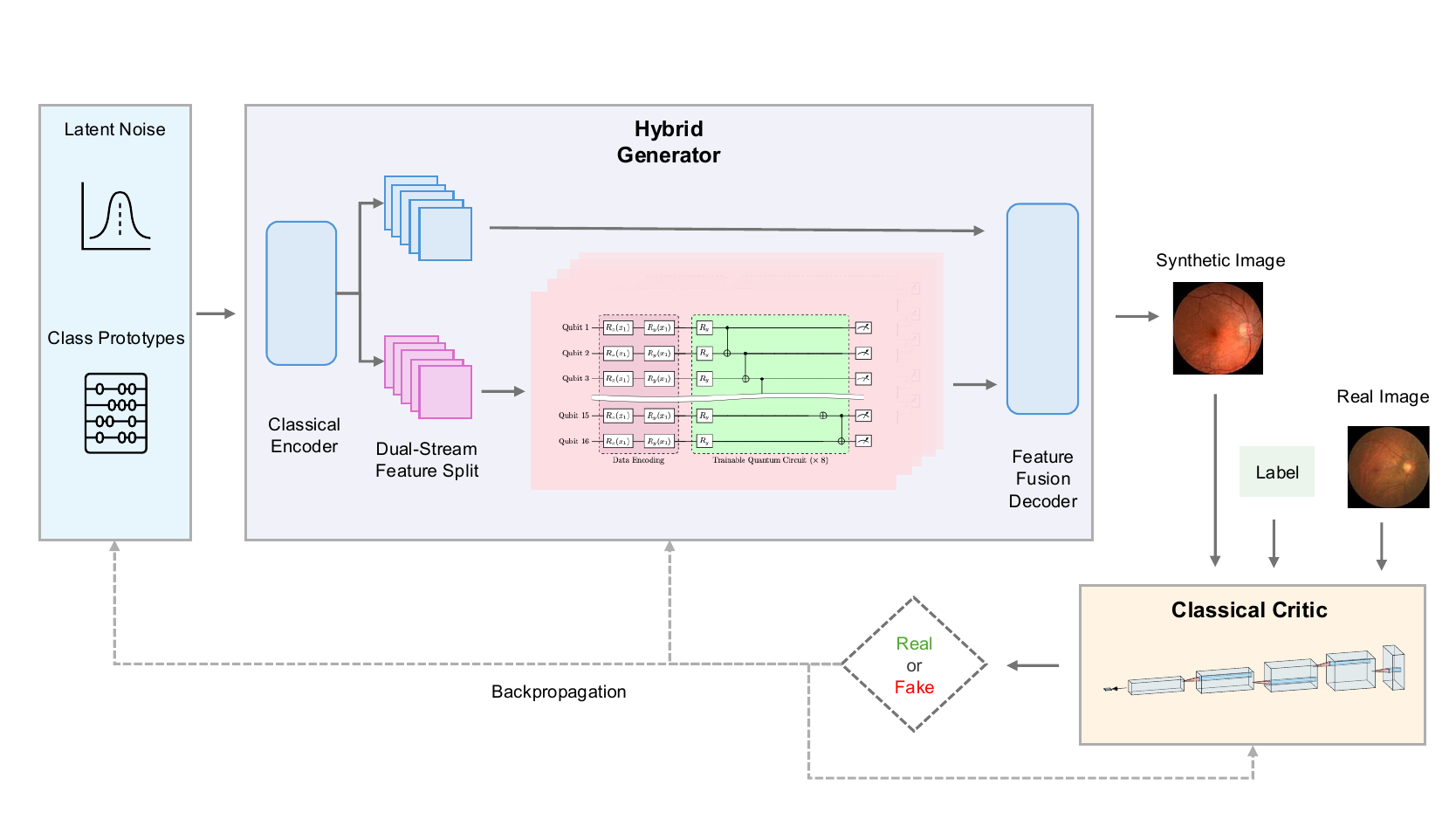}
    \caption{}
    \label{fig:gan_arch_a}
  \end{subfigure}

  \vspace{3mm} 

  \begin{subfigure}{\textwidth}
    \centering
    \includegraphics[width=0.62\linewidth]{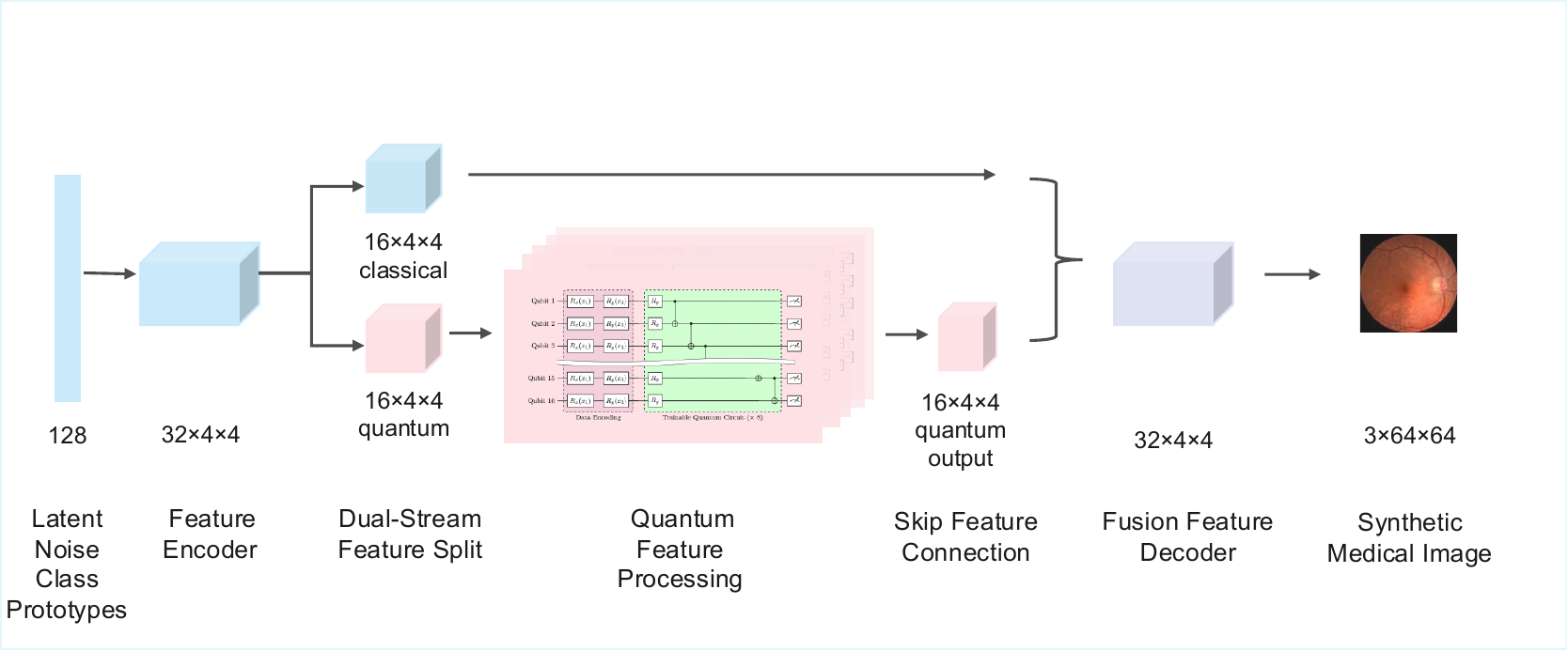}
    \caption{}
    \label{fig:gan_arch_b}
  \end{subfigure}

  \caption{\textbf{MediQ-GAN architecture.}
  \textbf{(a)} Full GAN pipeline. In MediQ-GAN, the latent noise is sampled from Gaussian noise and passed to a simple classical neural network. Half of the classical intermediate features are encoded into quantum features and processed by variational quantum circuits. The output from the quantum circuits is concatenated with the rest of the classical intermediate features, followed by a convolutional layer to produce the color image. The critic is a classical deep convolutional network and acts as a quality inspector, labeling input images as real or fake.
  \textbf{(b)} Generator architecture. The input $z$ (latent noise plus prototypes) is encoded into a feature map $F\in\mathbb{R}^{32\times4\times4}$, split into classical and quantum streams. Quantum features are processed by five variational quantum circuits (16 qubits each), then fused with classical features and decoded into a $3\times64\times64$ RGB image.}
  \label{fig:gan_arch}
\end{figure}

\subsection*{Overview}\label{result_overview}
We conduct experiments on three medical imaging benchmarks with limited data: ISIC 2019 for dermoscopy, ODIR-5k for fundus photography, and RetinaMNIST for downsampled fundus images, all at a resolution of $64 \times 64$. The size of each dataset is shown in Extended Data \ref{tab:dataset-size}. Our generator employs five quantum sub-generators with eight-layer depth, combined with classical skip-connected features, while the critic adopts a WGAN-GP architecture.

Our primary evaluation focuses on the practical utility of the proposed MediQ-GAN as a data augmentation tool for medical image analysis. We assess whether synthetic images generated by our model improve downstream classifier performance when added to the limited real training sets. Specifically, we evaluate classification accuracy (ACC) and the area under the ROC curve (AUC) on held-out test sets after augmentation. Accuracy provides a measure of overall performance, while AUC is robust to class imbalance, therefore suitable for assessing clinical relevance in datasets where rare but critical conditions must be captured.

To interpret the augmentation performance, we further evaluate the fidelity and diversity of generated images by Fréchet Inception Distance (FID)~\cite{10.5555/3295222.3295408} and intra-class Learned Perceptual Image Patch Similarity (LPIPS)~\cite{8578166}. FID captures the distributional discrepancy between real and generated images in a deep feature space, whereas LPIPS quantifies perceptual diversity within each class, revealing whether the model avoids mode collapse and generates a rich set of images.

We compare our approach against a diverse set of generative models, including both state-of-the-art GAN and diffusion-based models. Classical GAN baselines include DCGAN~\cite{radford2016unsupervisedrepresentationlearningdeep}, the fundamental deep convolutional GAN architecture, and WGAN-GP \cite{10.5555/3295222.3295327}, which we also adopt as the base training objective for our hybrid model due to its stabilization properties in hybrid GANs~\cite{Tsang_2023}. We further include Structure-Preserving GAN~\cite{DBLP:conf/icml/BirrellKRZ22}, designed to exploit intrinsic group symmetries, which has been shown to improve fidelity and diversity for medical images with limited samples. StyleGAN2 with Adaptive Discriminator Augmentation (ADA)~\cite{NEURIPS2020_8d30aa96} is included as a state-of-the-art solution for stabilizing GAN training on small datasets, and FastGAN~\cite{liu2021towards} as a lightweight architecture optimized for few-shot generation with minimal computational cost. All GANs are trained from scratch. For pretrained approaches, we compare against Diffusion Transformer (DiT)~\cite{10377858} and Stable Diffusion~3~\cite{10.5555/3692070.3692573}, both fine-tuned using parameter-efficient LoRA adapters~\cite{hu2021loralowrankadaptationlarge} to accommodate the limited data setting. 

Finally, we benchmark against MosaiQ~\cite{Silver_2023_ICCV}, the closest prior hybrid GAN. The original MosaiQ was designed for $28 \times 28$ grayscale images and employed eight quantum sub-generators with five qubits each. To adapt it for our higher-resolution, multi-class setting, we scaled the architecture to ten quantum sub-generators with ten qubits each and conditioned each sub-generator on class labels through separate label embeddings.

\subsection*{Downstream performance after augmentation}\label{result_classification}

\begin{table}[h]
\centering
\caption{\textbf{Classification performance comparison} of different methods for data augmentation across ISIC2019, ODIR-5k, and RetinaMNIST.}
\label{tab:all_aug}
\begin{tabular*}{\textwidth}{@{\extracolsep\fill}llrrrr}
\toprule
\textbf{Dataset} & \textbf{Method} & \multicolumn{2}{c}{EfficientNetB0} & \multicolumn{2}{c}{ViT-small} \\
\cmidrule(lr){3-4} \cmidrule(lr){5-6}
& & ACC(\%) & AUC & ACC(\%) & AUC \\
\midrule
\multirow{9}{*}{ISIC2019}
& Baseline     & 72.24 & 0.9230 & 72.49 & 0.9231 \\
& DCGAN        & 74.02 & 0.9316 & 78.48 & 0.9475 \\
& WGAN-GP      & 64.16 & 0.8444 & 66.44 & 0.8773 \\
& Structure-preserving GAN & 75.79 & 0.9313 & 79.31 & 0.9495 \\
& FastGAN      & 75.97 & 0.9386 & 80.22 & \textbf{0.9549} \\
& StyleGAN2-ADA& 74.86 & 0.9326 & 79.42 & 0.9519 \\
& DiT          & 73.82 & 0.9316 & 79.28 & 0.9516 \\
& Stable Diffusion 3 & 74.30 & 0.9269 & 79.26 & 0.9470 \\
& MosaiQ        & 62.73 & 0.9431 & 50.25 & 0.8477 \\
& \textbf{MediQ-GAN}         & \textbf{75.99} & \textbf{0.9386} & \textbf{82.60} & 0.9517 \\
\midrule
\multirow{9}{*}{ODIR-5k}
& Baseline     & 52.69 & 0.7907 & 55.62 & 0.8191 \\
& DCGAN        & 55.51 & 0.7941 & 56.52 & 0.8140 \\
& WGAN-GP      & 54.33 & 0.7920 & 54.41 & 0.7974 \\
& Structure-preserving GAN & 55.27 & 0.8055 & 55.43 & 0.8009 \\
& FastGAN      & 56.52 & 0.8101 & 58.63 & 0.8553 \\
& StyleGAN2-ADA& 53.87 & 0.7938 & 55.19 & 0.8201 \\
& DiT          & 58.40 & 0.8934 & 55.66 & 0.8084 \\
& Stable Diffusion 3 & 53.08 & 0.7917 & 53.71 & 0.7883 \\
& MosaiQ       & 50.13 & 0.7235 & 51.76 & 0.7767 \\
& \textbf{MediQ-GAN}          & \textbf{59.52} & \textbf{0.8997} & \textbf{59.06} & \textbf{0.8591} \\
\midrule
\multirow{9}{*}{RetinaMNIST}
& Baseline     & 47.50 & 0.7361 & 50.93 & 0.7740 \\
& DCGAN        & 51.25 & 0.7683 & 55.93 & 0.7941 \\
& WGAN-GP      & 54.37 & 0.7880 & 52.81 & 0.7932 \\
& Structure-preserving GAN & 43.75 & 0.6130 & 40.31 & 0.6828 \\
& FastGAN      & 53.75 & 0.8187 & \textbf{64.37} & 0.8920 \\
& StyleGAN2-ADA& 50.31 & 0.7061 & 50.12 & 0.7147 \\
& DiT          & 55.62 & 0.8041 & 53.12 & 0.7924 \\
& Stable Diffusion 3 & 47.31 & 0.7287 & 50.62 & 0.7386 \\
& MosaiQ         & 41.33 & 0.7104 & 47.81 & 0.7564 \\
& \textbf{MediQ-GAN}          & \textbf{56.75} & \textbf{0.8283} & 63.75 & \textbf{0.8975} \\
\bottomrule
\end{tabular*}
\end{table}

We augment training sets to class balance using each generator and fine-tune EfficientNet-B0~\cite{pmlr-v97-tan19a} and Vision Transformer (ViT-small)~\cite{DBLP:conf/iclr/DosovitskiyB0WZ21} as classifiers. MediQ-GAN model delivers the largest or near-largest gains in accuracy and AUC across datasets, as shown in Table \ref{tab:all_aug}.

\textbf{ISIC 2019:}
MediQ-GAN achieves the highest EfficientNet-B0 accuracy at 75.99\%, improving over the baseline by 3.75 \%, with AUC rising from 0.9230 to 0.9386. 
With ViT-small, MediQ-GAN improves accuracy from 72.49\% to 82.60\% and AUC from 0.9231 to 0.9517. Although FastGAN yields a slightly higher ViT AUC of 0.9549, our model achieves the largest gain in accuracy. In contrast, MosaiQ underperforms significantly, with ViT-small accuracy dropping to 50.25\%. Its generated samples appear as blurred color patches lacking meaningful high-resolution structures, which severely limits their usefulness for classifier training.

\textbf{ODIR-5k:}
Our method achieves the best results across both backbones. EfficientNet-B0 accuracy rises from 52.69\% to 59.52\%, while AUC improves from 0.7907 to 0.8997. ViT-small accuracy improves from 55.62\% to 59.06\% and AUC from 0.8191 to 0.8591. In contrast, DiT produces strong AUC gains on EfficientNet-B0 but underperforms on ViT.

\textbf{RetinaMNIST:}
In the most data-scarce setting, our model attains 56.75\% accuracy and 0.8283 AUC with EfficientNet-B0, and 63.75\% accuracy with 0.8975 AUC on ViT-small. These correspond to improvements of 9.25\% and 12.82\% in accuracy over the baselines. The performance boost surpasses both FastGAN and DiT and validates the effectiveness of our generator for few-shot augmentation.

Across all three datasets, MediQ-GAN consistently improves both accuracy and AUC over baselines. On ISIC 2019, ODIR-5k, and RetinaMNIST, the average accuracy increased by 3.75 \%, 6.83 \%, and 9.25 \%, and the average AUC improved by 0.016, 0.109, and 0.092, respectively. These results demonstrate that data augmentation using our method can significantly improve the performance of medical image classification tasks. 

\subsection*{Generated image quality and diversity}

\begin{figure}[t]
  \centering
  \includegraphics[width=\textwidth]{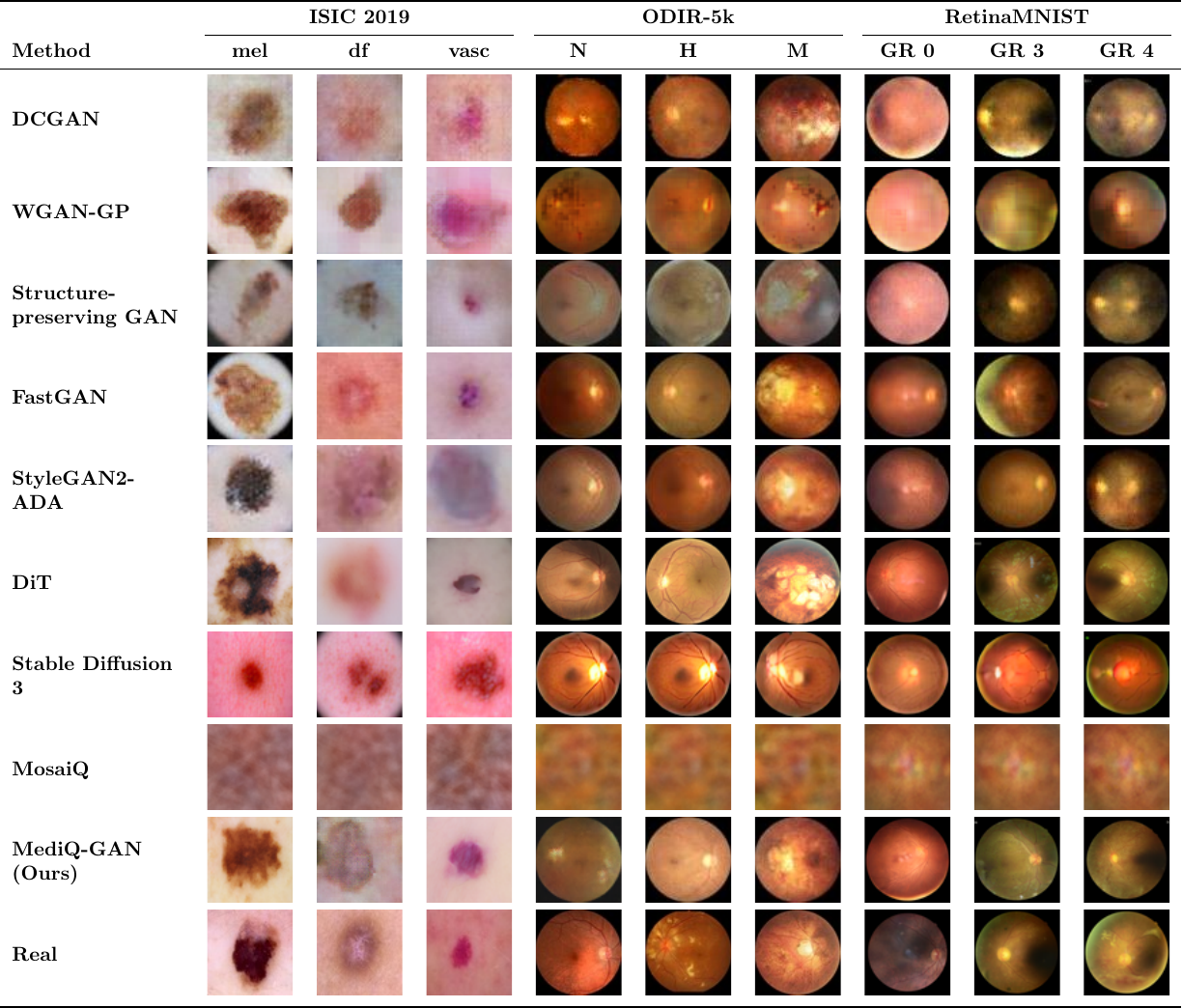}
  \caption{\textbf{Random synthetic samples} from representative classes of ISIC 2019, ODIR-5k, and RetinaMNIST by baseline generative models and our model, MediQ-GAN.}
  \label{fig:generated_images_grid}
\end{figure}

\setlength{\tabcolsep}{3pt} 
\begin{table*}[t]
\caption{\textbf{Per-class FID ($\downarrow$) scores} of representative classes of ISIC 2019, ODIR-5k, and RetinaMNIST.}
\label{tab:fid-multi}
\centering
\resizebox{\textwidth}{!}{%
\begin{tabular}{@{}l
   rrr 
   rrr 
   rrr 
   @{}}
\toprule
& \multicolumn{3}{c}{\textbf{ISIC 2019}} 
& \multicolumn{3}{c}{\textbf{ODIR-5k}} 
& \multicolumn{3}{c}{\textbf{RetinaMNIST}} \\
\cmidrule(lr){2-4}\cmidrule(lr){5-7}\cmidrule(lr){8-10}
\textbf{Method} 
& \textbf{mel} & \textbf{df} & \textbf{vasc}
& \textbf{N}   & \textbf{H}  & \textbf{M}
& \textbf{GR 0} & \textbf{GR 3} & \textbf{GR 4} \\
\midrule
DCGAN      & 230.8 & 272.6 & 228.8 & 226.9 & 235.1 & 212.9 & 216.0 & 181.8 & 192.9 \\
WGAN-GP    & 185.3 & 309.3 & 307.6 & 201.6 & 212.7 & 193.9 & 153.2 & 164.1 & 177.3 \\
Structure\mbox{-}preserving GAN 
           & 233.3 & 324.6 & 285.4 & 241.3 & 276.2 & 246.9 & 235.7 & 242.7 & 442.5 \\
FastGAN    & 212.5 & 225.0 & 225.0 & 220.5 & 242.5 & 197.6 & 193.7 & 175.2 & 181.3 \\
StyleGAN2-ADA 
           & 231.5 & 239.5 & 401.7 & 237.7 & 255.8 & 205.6 & 211.8 & 179.0 & 252.3 \\
DiT        & 189.1 & 200.2 & 174.7 & \textbf{153.5} & 200.0 & 149.4 & \textbf{122.0} & 113.9 & 123.4 \\
Stable Diffusion 3 
           & 245.7 & 191.2 & 235.7 & 210.7 & 219.2 & 211.7 & 164.7 & 161.8 & 160.2 \\
MosaiQ     & 506.9 & 440.8 & 435.0 & 552.7 & 547.8 & 593.2 & 382.4 & 343.3 & 316.5 \\
\textbf{MediQ-GAN} 
           & \textbf{187.3} & \textbf{182.2} & \textbf{171.4} 
           & 162.8 & \textbf{167.4} & \textbf{144.3} 
           & 127.3 & \textbf{159.7} & \textbf{116.9} \\
\bottomrule
\end{tabular}%
}
\end{table*}

To understand the gains observed in downstream classification, we next evaluate the fidelity and diversity of generated images. Across all datasets, MediQ-GAN consistently achieves the best or second-best per-class FID (Table~\ref{tab:fid-multi}). On ISIC 2019, MediQ-GAN outperforms all classical baselines on rare classes (df, vasc, and scc). Even on classes with more training samples, our approach surpasses FastGAN and StyleGAN2-ADA and matches DiT, despite DiT's substantially larger capacity. On ODIR-5k, our method performs particularly well on minority classes (G, A, H, M), where oversmoothing and loss of fine vascular detail are common failure modes for other GANs. On RetinaMNIST, our model achieves the lowest FID in 4 of 5 classes, demonstrating robust performance in few-shot regimes. The averages of the majority and minority class metrics are in Extended Data \ref{tab:fid-minor-major}. Full per-class metrics and samples are provided in the Supplementary Information.

Besides fidelity, we evaluate intra-class diversity using LPIPS in Extended Data \ref{tab:lpips-minor-major}. While DCGAN and WGAN-GP can achieve diversity scores comparable to real data when there are more training samples, their high FID values indicate that this diversity largely reflects noise rather than true coverage of the data manifold. Under limited-data conditions, most clearly in RetinaMNIST, baselines such as StyleGAN2-ADA and Stable Diffusion 3 exhibit severe mode collapse, producing visually similar samples or missing class-specific features. FastGAN, DiT, and our model remain robust, with our approach consistently ranking closest or second-closest to real images in intra-class LPIPS while achieving competitive FID. These combined quantitative results demonstrate that our approach generates images that are both high-fidelity and diverse across all classes.

Despite our architectural and training modifications, MosaiQ remains constrained by its PCA-based dimensionality reduction, which leads to severe loss of color fidelity and structural detail in the generated images. The resulting samples are blurred and lack realistic class-specific features, and its FID scores are consistently higher than those of both the classical baselines and our model. These results indicate that MosaiQ does not scale effectively to high-resolution color image generation, highlighting the necessity of our design.

In comparison, the other baselines exhibit a range of drawbacks that further highlight the strengths of our approach. Qualitative comparisons (Fig.~\ref{fig:generated_images_grid}) reinforce these findings. DCGAN and Structure-Preserving GAN frequently produce blurred dermoscopic textures, while StyleGAN2-ADA improves majority-class quality but struggles with scarce classes. FastGAN and DiT perform well with sufficient or moderately limited data, but lose structure under extreme scarcity (class df in ISIC). Stable Diffusion 3 generates visually plausible vessels but exhibits very low inter-class variability. In contrast, MediQ-GAN preserves class-specific lesion cues (ISIC 2019) and vascular morphology (ODIR-5k/RetinaMNIST) even under data scarcity. Therefore, MediQ-GAN demonstrates its utility as a reliable augmentation tool for training robust classifiers. This balance of fidelity and diversity is critical for improving downstream diagnostic models, particularly in detecting rare diseases where limited real-world data are available.

\subsection*{NISQ hardware evaluation}
Our approach is quantum-inspired and hardware-agnostic: the correctness does not depend on quantum execution, and quantum devices serve only to accelerate computation. To validate robustness under realistic conditions, we executed the five 16-qubit VQCs on the ibm\_marrakesh processor. Due to queue times and shot limits, we restricted hardware tests to inference of representative classes of each dataset. The results are shown in Extended Data~\ref{tab:fid-real}. As expected, performance on real hardware was slightly below noiseless simulation but remained comparable to simulated results and strong classical baselines. This result proves the robustness and practical viability of our quantum-inspired design under NISQ noise, while our main contributions are independent of specific quantum hardware.

\section*{Discussion}\label{discussion}
In this paper, we presented a novel quantum-inspired generative adversarial network, MediQ-GAN, for high-resolution medical image generation. Our approach leverages the complementary strengths of classical deep learning and quantum computing. Classical layers efficiently capture global structures, while quantum sub-generators enhance expressivity in high-dimensional Hilbert space. By incorporating dynamic prototypes and skip-feature connections that fuse classical and quantum feature maps, our model captures both coarse and fine-grained phenotypes in limited-data regimes. Empirically, we found that MediQ-GAN produces images with higher fidelity and diversity compared with state-of-the-art generative models, demonstrating its ability to generate clinically meaningful samples that augment underrepresented classes. Beyond performance, we conducted a careful analysis of the trainability and expressivity of the hybrid generator in the Method section \ref{method_rank_analysis}. The Jacobian singular value spectrum and effective rank show that the decay of singular values is smoother than in classical baselines, indicating that a greater proportion of latent directions meaningfully influence the output \cite{pmlr-v80-odena18a, DBLP:conf/iclr/WangP21}. This richer utilization of the latent space suggests that the hybrid generator avoids mode collapse and encourages diversity. In addition, our dynamical Lie algebra (DLA) analysis confirmed that the chosen quantum circuit ansatz at its optimal depth achieves sufficient expressivity while safely avoiding barren plateaus \cite{Ragone_2024, Fontana_2024}. Finally, we demonstrated the real-world applicability of our model by using the generated images for data augmentation on class-imbalanced medical imaging datasets. The resulting improvements in test accuracy and AUC scores consistently exceeded those achieved with classical model-based augmentation. 

Future studies on quantum-enhanced GANs can focus on several aspects to further improve training efficiency, performance, scalability, and generalizability. First, while the current work employed a hardware-efficient, problem-agnostic ansatz, tailoring circuit architectures to image generation tasks could further improve expressivity-to-trainability trade-offs, potentially reducing circuit depth without sacrificing model capacity. Second, incorporating quantum-aware optimization methods, such as Quantum Natural Gradient (QNG) descent \cite{Stokes_2020}, could improve the effective conditioning of the parameter space, leading to smoother loss landscapes and faster convergence. By leveraging the quantum Fisher information matrix as a natural metric, QNG adapts parameter updates to the geometry of the variational quantum circuit, potentially mitigating barren plateaus and enhancing trainability. Third, benchmarking can be extended to additional imaging modalities, such as histopathology slides, chest X-rays, CT, and MRI scans, as well as to non-medical but similarly data-scarce domains, such as satellite imagery. Such a broader evaluation would help uncover domain-specific challenges and provide insights to enhance the robustness and generalizability of the proposed hybrid GAN.

In conclusion, our work demonstrates the promise of hybrid classical–quantum GANs in addressing real-world challenges in data-scarce medical imaging. While MediQ-GAN has not yet surpassed the classical generative models in all unconstrained benchmarks, rapid advances in quantum error correction and scalability indicate strong future potential for fast, efficient, and high-quality image generation in limited-data settings. Our findings provide both a practical implementation and a theoretical framework that can guide the development of next-generation quantum-inspired generative models.

\section*{Methods}\label{methods}
In this section, we first provide a concise overview of classical GAN architectures and existing quantum-enhanced GAN approaches. We then describe our proposed GAN architecture and justify the trade-off between expressivity and trainability in the quantum ansatz. We also present an analysis of the hybrid generator’s performance based on the Jacobian singular value spectrum and effective rank, providing a principled comparison with classical baselines. Details of the experimental setup and hyperparameters are provided in the Supplementary Information.

\subsection*{Background: classical and quantum generative adversarial networks}\label{method_background}
Generative Adversarial Networks (GANs) have emerged as a powerful class of deep generative models, defining data generation as a minimax game between a generator and a discriminator~\cite{Goodfellow2014}. While effective, this original formulation often suffered from vanishing gradients and mode collapse.

To address these limitations, Arjovsky et al.~\cite{Arjovsky2017} proposed the Wasserstein GAN (WGAN), which replaces the Jensen–Shannon divergence implicit in the original GAN with the Earth Mover’s (Wasserstein-1) distance.  Let \(p_{\text{data}}\) denote the data distribution, \(p_z\) the latent noise distribution, \(G(z)\) the generator, and \(D(\cdot)\) the 1-Lipschitz critic. The WGAN objective is given as:
\begin{equation}\label{eq:wgan-loss}
\mathcal{L}_{\text{WGAN}}(G,D) =
\mathbb{E}_{x \sim p_{\text{data}}}[D(x)] -
\mathbb{E}_{z \sim p_z}[D(G(z))].
\end{equation}

Here
$D:\mathcal{X}\to\mathbb{R}$
is a real-valued score function. The critic maximizes
$\mathcal{L}_{\text{WGAN}}$
while the generator minimizes it, or equivalently maximizing
$\mathbb{E}_{z}[D(G(z))].$ This modification ensures that the loss is continuous and provides informative gradients even when the generator distribution and data distribution have little overlap, thus significantly improving training stability. In practice, enforcing the Lipschitz constraint via weight clipping proved suboptimal and often led to capacity underutilization. Gulrajani et al.~\cite{Gulrajani2017} introduced WGAN with Gradient Penalty (WGAN-GP), which instead enforces the Lipschitz condition by penalizing the norm of the critic’s gradient:

\begin{equation}
\lambda \, \mathbb{E}_{\hat{x} \sim p_{\hat{x}}}\Big[
  \big(\|\nabla_{\hat{x}} D(\hat{x})\|_2 - 1 \big)^2
\Big],
\end{equation}

where $\hat{x}$ is sampled uniformly along straight lines between real and generated samples. Despite their success, state-of-the-art classical models often require substantial computational resources and struggle with the scarcity of data and the class imbalance inherent in specialized domains such as medical imaging~\cite{FridAdar2018, Sandfort2019}. 

Quantum machine learning (QML) offers a promising way to address these limitations by leveraging uniquely quantum resources such as superposition and entanglement to compactly represent complex probability distributions. Variational quantum circuits (VQCs) provide a trainable framework that can be optimized in a hybrid quantum-classical loop, employing classical loss functions and optimizers for quantum parameterized gates. In principle, this enables highly expressive models with potentially fewer parameters or improved generalization in low-data regimes.

Quantum GANs (QGANs) were first introduced theoretically~\cite{Lloyd2018, DallaireDemers2018} and have since been demonstrated on noisy intermediate-scale quantum (NISQ) devices~\cite{Hu2019, Huang2021}. Recent hybrid models, such as QGPatch~\cite{Huang2021} and MosaiQ~\cite{Silver_2023_ICCV}, combine quantum sub-generators with classical preprocessing or dimensionality reduction networks. While promising, these approaches remain constrained to generating very low-resolution grayscale images (e.g., $8 \times 8$ or $28 \times 28$) and often rely on aggressive dimensionality reduction that discards potentially informative features. Tsang et al.~\cite{Tsang_2023} recently proposed a hybrid quantum-classical GAN framework for image generation without dimensionality reduction, achieving comparable performance to classical GANs with far fewer generator parameters. Their work systematically investigates the effects of circuit width, patch size, and latent priors, providing valuable insights into hybrid model design. However, their experiments remain limited to grayscale images at $28 \times 28$ resolution, and do not include formal analyses of generator trainability and expressivity. 

In contrast to previous hybrid quantum–classical approaches, we adopt a quantum-inspired perspective, leveraging the inherent full-rank transformations of quantum circuits. In this view, quantum devices serve mainly to accelerate execution rather than introduce fundamentally new functionality, which distinguishes our method from prior works constrained by NISQ hardware. This hardware-agnostic design avoids near-term constraints, scales to high-resolution color images, and enables rigorous analyses of trainability and expressivity.

\subsection*{Quantum-inspired GAN}\label{method_hybrid_arch}
\subsubsection*{Prototype-based conditioning}
Prototype-based conditioning has been shown to stabilize GAN training, improve representation learning, and mitigate class imbalance in limited-data settings~\cite{li2022prototype, 9022182}. By summarizing class distributions into representative embeddings, prototypes act as low-variance anchors that guide the generator toward essential phenotype features, reducing overfitting to noisy samples and mitigating mode collapse for rare classes. 

In our framework, we fine-tune an EfficientNetB0 backbone for 30 epochs and extract the final-layer activations as features. A single averaged centroid is computed for each class, and this prototype vector is concatenated with the latent noise vector to form the input to the hybrid generator. Intuitively, this provides a coarse semantic “starting point” for generation, allowing the network to focus its capacity on learning intra-class variations rather than rediscovering the global class structure from scratch. 

A common challenge with prototype conditioning is reduced diversity, since naive conditioning can bias the generator toward a single prototype representation per class. We address this by (i) leveraging the high-dimensional feature space of the quantum sub-generators, which enrich intra-class variability, and (ii) introducing a prototype regularization loss that explicitly encourages inter-class separation by maximizing the pairwise distances between class centroids. This combination yields both diversity within each class and discrimination between classes, reducing the risk of prototype collapse.

\subsubsection*{Hybrid generator}
The hybrid generator architecture is illustrated in Figure~\ref{fig:gan_arch_b}. We first map the class-conditioned latent $z\!\in\!\mathbb{R}^{128}$ into an initial feature map with a lightweight convolutional encoder. Specifically, $z$ is reshaped to $(B,128,1,1)$, upsampled to $4{\times}4$, and passed through $1{\times}1$ and $3{\times}3$ convolutions to produce a shared feature map $F\in\mathbb{R}^{32\times 4\times 4}$.

We then form two parallel streams. The quantum stream receives half of the shared $4{\times}4$ features, each of which is flattened, reshaped, and then processed by an independent 16-qubit variational quantum circuit (VQC) with $L$ layers of entangling gates. Each VQC outputs a 16-dimensional vector, which is reshaped back to $4{\times}4{\times}1$ and concatenated to form $F_{\text{q}}\!\in\!\mathbb{R}^{16\times4\times4}$. In parallel, the classical stream produces $F_{\text{cls}}\!\in\!\mathbb{R}^{16\times4\times4}$, which provides complementary global context. The fused representation $F_{\text{fuse}}\!\in\!\mathbb{R}^{32\times4\times4}$ is passed through a decoder consisting of a $3{\times}3$ convolution followed by three stages of upsample–blur–convolution blocks \cite{Karras2020StyleGAN2} and a final $3{\times}3$ convolution, progressively generating a $3\times64\times64$ RGB image. The training hyperparameters are described in the Supplementary Information.

This dual-stream design with skip connections is similar to multi-branch generators in BigGAN\cite{brock2018large} and StyleGAN \cite{Karras2020StyleGAN2}, where skip pathways help maintain global structure and the other branches refine high-frequency details. By independently varying the feature split proportion and the number of quantum sub-generators, we systematically analyze how the quantum circuits affect downstream classification performance, synthetic image fidelity, and effective rank. The corresponding ablation results are provided in the Supplementary Information. These experiments show that allocating more quantum feature channels or adding additional quantum sub-generators generally improves performance and effective rank, but the marginal gains diminish beyond the current configuration. To balance simulation cost and available quantum resources, we adopt a configuration of 16 quantum channels and five VQCs. As quantum hardware becomes more capable, scaling up the quantum branch would be a natural next step to further enhance performance.

\subsubsection*{Classical critic}
\label{method_critic}
For the discriminator (critic), we adopt the convolutional architecture from the original WGAN-GP paper~\cite{10.5555/3295222.3295327}. Retaining a purely classical critic is motivated by both practicality and training dynamics. 

From a hardware perspective, the critic must process high-dimensional $3\times 64 \times 64$ image inputs at every training iteration. Implementing such a network on current NISQ devices would be prohibitively expensive, whereas classical convolutional networks can perform this task efficiently. 

From an optimization perspective, the critic serves as the “adversarial signal generator” that drives the learning process of the hybrid generator. Using a stable and well-characterized classical critic ensures that improvements in generator performance can be attributed to the quantum-inspired architecture rather than potential instabilities or noise from a quantum critic. This design choice aligns with most prior hybrid quantum--classical GAN works~\cite{Silver_2023_ICCV, Tsang_2023}, where the generator is hybrid classical and quantum while the critic remains purely classical.

\subsubsection*{Variational circuit ansatz}
\begin{figure}[t]
  \centering
  \includegraphics[width=\linewidth]{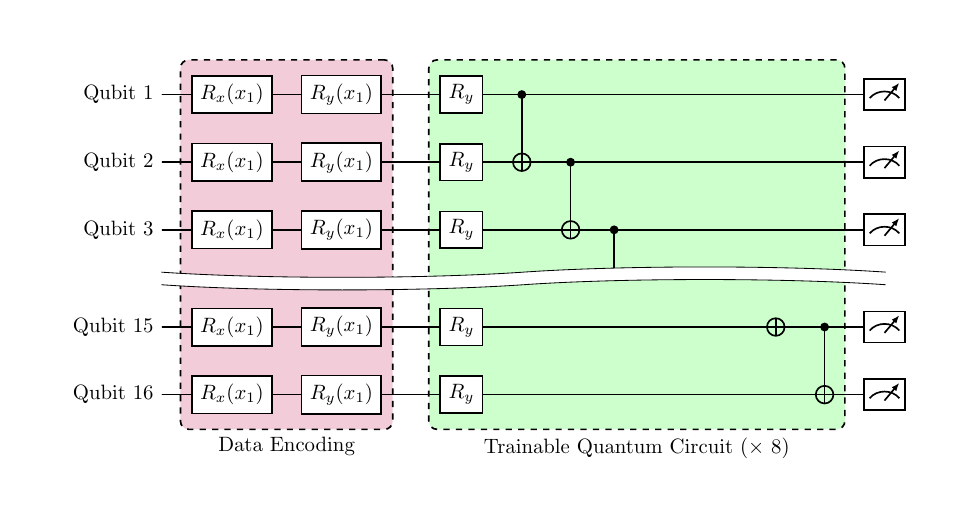} 
  \caption{\textbf{Variational quantum circuit ansatz design for quantum generators.} Each input component \(x_j\) is encoded with angle encoding to qubit \(j\) using \(R_x(x_j)\) followed by \(R_y(x_j)\). The encoded state is processed by \(L\) identical trainable layers ($L=8$ for optimal result), each applying a learnable single-qubit rotation \(R_y(\theta_{\ell,j})\) and a nearest-neighbour CNOT (\(1\!\to\!2\!\to\!\cdots\!\to\!N\)). All qubits are finally read out in the \(X\) basis and produce the classical vector \(\mathbf{m}(x;\Theta)=(\langle X_1\rangle,\ldots,\langle X_N\rangle)\in[-1,1]^N\). The “\(\times 8\)” indicates repetition layers, and the wavy gap indicates middle wires with the same gates.}
  \label{fig:vqc_graph}
\end{figure}

We employ a hardware-efficient variational quantum circuit (VQC) with two stages, as illustrated in Figure~\ref{fig:vqc_graph}. The design follows the structure of MosaiQ~\cite{Silver_2023_ICCV}, which has been shown to be simple yet expressive, and is suitable for noisy intermediate-scale quantum (NISQ) devices. 

Starting from the all-zero state $\lvert 0 \rangle^{\otimes N}$, the classical input vector $z \in \mathbb{R}^N$ is encoded into a quantum state via angle encoding, where each normalized feature value determines the rotation angles about the $X$- and $Y$-axes:
\begin{equation}
W(z) = \prod_{j=1}^{N} R_y(\beta x_j) R_x(\alpha x_j),
\qquad
R_x(\phi)=e^{-i\phi X/2}, \quad R_y(\theta)=e^{-i\theta Y/2}
\end{equation}

This forms the data-encoding block of the circuit. 

The encoded state is then processed by a trainable variational block, repeated $L$ times, where each layer applies parameterized single-qubit $R_y$ rotations followed by nearest-neighbor CNOT gates to introduce entanglement:
\begin{equation}
U^{(\ell)}(\Theta^{(\ell)})=\Bigg(\prod_{j=1}^{N-1}\mathrm{CNOT}_{j\rightarrow j+1}\Bigg)
\Bigg(\prod_{j=1}^{N} R_y(\theta^{(\ell)}_{j})\Bigg)
\end{equation}

The full ansatz is given by
\begin{equation}
{\lvert \psi \rangle(z;\Theta)}=\Bigg(\prod_{\ell=1}^{L} U^{(\ell)}(\Theta^{(\ell)})\Bigg)\,W(z)\,\lvert{0}\rangle^{\otimes N}
\end{equation}

where $\Theta=\{\theta^{(\ell)}_{j}\}$ are the trainable parameters, totaling $N \times L$. Each qubit is measured in the $X$ basis, producing the classical readout vector $\mathbf{m}(z;\Theta)\in[-1,1]^N$. To ensure numerical stability and smooth integration with the downstream classical network, we apply a lightweight normalization–nonlinearity head:
\begin{equation}
h(\mathbf{m}) = \sigma_{\mathrm{LeakyReLU}}\!\big(\mathrm{LN}(\mathbf{m})\big)
\end{equation}

where $\mathrm{LN}$ denotes Layer Normalization and the LeakyReLU slope $\alpha=0.2$.

This ansatz balances expressive power with hardware efficiency, and its layer depth $L$ governs the trade-off between representation capacity and susceptibility to barren plateaus. In the next subsection, we analyze its trainability and expressivity using Jacobian singular value spectra and dynamical Lie algebra (DLA) theory. Results of the layer-depth sensitivity study are provided in the Supplementary Material.

\subsection*{Latent Geometry and Performance Analysis}\label{method_rank_analysis}
\begin{figure*}[t]
  \centering
  \includegraphics[width=\textwidth]{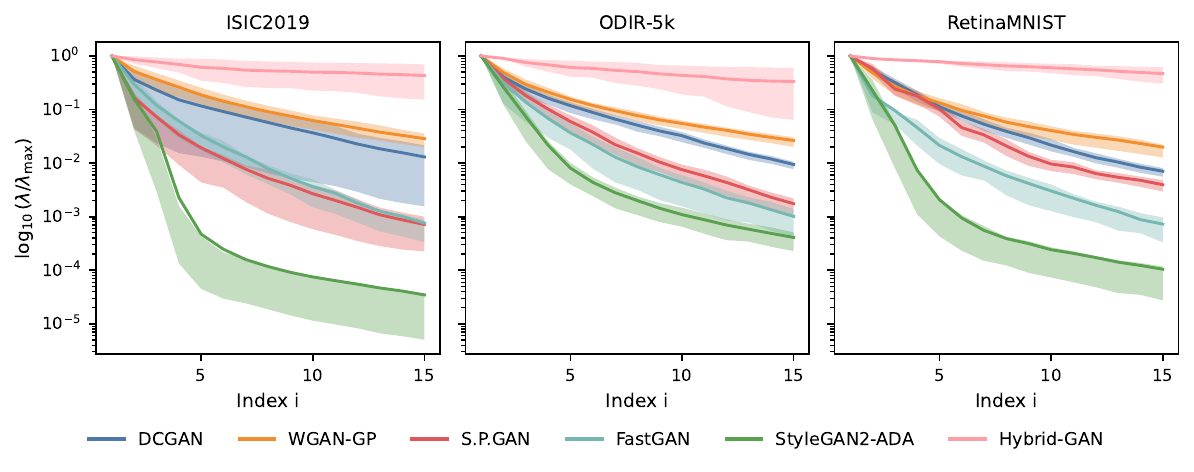}
  \caption{\textbf{Latent–Jacobian spectra.} Normalized squared singular values $\lambda_i/\lambda_\text{max}$ of the generator Jacobian $J(z)$ are plotted on a $\log_{10}$ scale. Solid lines indicate the mean across runs, and shaded regions represent the 25th-75th percentiles. Flatter spectra indicate higher effective rank, while steeper spectra indicate concentration into a few directions.}
  \label{fig:jacobian-spectra}
\end{figure*}

Although quantum-enhanced GANs often achieve superior performance with fewer parameters, a formal explanation of these gains remains unclear. Inspired by \cite{pmlr-v80-odena18a} and \cite{DBLP:conf/iclr/WangP21}, we adopt a geometry-based approach to analyze GAN performance. These works show that the singular value spectrum of the generator Jacobian encodes how effectively a model utilizes its latent space, with flatter spectra and higher effective ranks associated with greater diversity and better generalization.

Following this insight, we compute the Jacobian
$J(z)=\partial G(z)/\partial z\in\mathbb{R}^{n\times d_z}$, where $d_z$ is the latent input dimension and $n=3HW$ for flattened RGB images. The Jacobian maps latent noises to output images.
The singular value decomposition $J=U\Sigma V^\top$ yields singular values
$\{\sigma_i\}_{i=1}^{d_z}$, whose squares
$\lambda_i=\sigma_i^2$ are the eigenvalues of $J^\top J$ and quantify local sensitivity along orthogonal latent directions. The shape of the spectrum $\{\lambda_i\}$ reflects both the expressivity and the numerical conditioning of the generator mapping \cite{pmlr-v80-odena18a,DBLP:conf/iclr/WangP21}. 

We summarize this spectrum with three metrics.
First, the effective rank
\begin{equation}
\mathrm{erank}_{\sigma^2}=\exp\!\Bigl(-\textstyle\sum_{i=1}^{d_z} p_i \log p_i\Bigr),
\qquad
p_i=\frac{\lambda_i}{\sum_{j=1}^{d_z}\lambda_j}
\end{equation}

uses the Shannon entropy of the normalized squared singular values to estimate the effective number of active latent directions.
Second, the participation ratio (PR)
\begin{equation}
\mathrm{PR}=\frac{\bigl(\sum_{i=1}^{d_z}\lambda_i\bigr)^2}{\sum_{i=1}^{d_z}\lambda_i^2}
\end{equation}

increases with spectral uniformity. Higher PR values indicate more uniform usage of latent directions.
Finally, the spectral condition number
\begin{equation}
\kappa_\varepsilon(J)=\frac{\sigma_{\max}}{\max(\sigma_{\min},\varepsilon)}
\end{equation}

captures local anisotropy. Lower values indicate a better-conditioned, more isotropic mapping.

In our experiments, we estimate $J$ via central finite differences with step size $\epsilon$ on random latent inputs for each model and dataset. We report the normalized versions over input latent dimension $d_z$: $\widehat{\mathrm{erank}}=\mathrm{erank}/d_z$ and $\widehat{\mathrm{PR}}=\mathrm{PR}/d_z$ to enable comparison across architectures. Across all datasets, MediQ-GAN exhibits a flatter singular-value spectrum (Fig.~\ref{fig:jacobian-spectra}), higher $\widehat{\mathrm{erank}}$ and $\widehat{\mathrm{PR}}$ (Extended Data~\ref{fig:erank_pr}), and the lowest $\log_{10}\kappa_\varepsilon(J)$ among baselines (Extended Data~\ref{tab:erank_multi}). Full quantitative results for each dataset are provided in the Supplementary Information. This indicates that MediQ-GAN activates more latent degrees of freedom and distributes sensitivity more evenly, which are critical properties for improved trainability and reduced mode collapse \cite{pmlr-v80-odena18a,DBLP:conf/iclr/WangP21} and align with its superior FID, LPIPS, and downstream classification performance. 

Prior work on classical GANs shows that deep generators without explicit spectral control can develop ill-conditioned Jacobians and concentrate sensitivity into a few directions. This dimensional concentration degrades sample quality and stability. Improving the input–output Jacobian and its singular-value distribution improves optimization, while poor conditioning correlates with worse Inception Score and FID \cite{pmlr-v80-odena18a}. State-of-the-art GAN training therefore introduces spectral and path-length regularizers that equalize the generator’s Jacobian and mitigate collapse of sensitivity to a low-dimensional subspace \cite{Karras2020StyleGAN2}. Our measurements are consistent with these findings. FastGAN and StyleGAN2 achieve higher effective rank and participation ratio than simpler GANs; however, StyleGAN2 still shows larger condition numbers, indicating ill-conditioning and a highly anisotropic latent space.

Quantum blocks provide a potential mechanism for higher effective rank and smoother spectra. Parameterized quantum circuits generate features in a Hilbert space whose dimension scales exponentially with qubit count. Entangling gates spread local sensitivity over many directions, which increases the number of influential latent directions and improves $\mathrm{erank}$ and PR \cite{Sim_2019,Havl_ek_2019}. Expectation values of bounded observables under unitary evolution impose Lipschitz control on the feature maps, which avoids extreme amplification along a few directions and yields smoother, less spiky singular-value spectra \cite{Sweke_2020,PhysRevResearch.6.043326}. Moreover, our ansatz exhibits a high participation ratio of the Quantum Fisher Information Matrix spectrum and a smooth eigenvalue decay. As detailed in Section~\ref{sec:theory}, the architecture avoids barren plateaus and preserves gradient signals, therefore keeping the sensitivity spectrum well spread. Together, these properties encourage isotropic use of the latent space and help explain the observed performance gains.

\subsection*{Theoretical framework for a trainable PQC architecture}\label{sec:theory}

In this section, we provide an analysis for our Parametrized Quantum Circuit (PQC)'s trainability and expressivity trade-off. The trainability and expressivity of a Parametrized Quantum Circuit (PQC) are intrinsically linked through its underlying algebraic and causal structure. For our linear-chain architecture, this relationship is governed by the principle of local interactions, which introduces a fundamental trade-off. While locality restricts expressivity to mitigate barren plateaus, it also mandates a minimum circuit depth to achieve system-wide entanglement. Therefore, we now formalize this balance using dynamical Lie algebra (DLA) and light-cone analysis to derive a principled guideline for optimal circuit depth.

\subsubsection*{Algebraic structure and barren plateau avoidance}

A key failure mode in training PQCs is the barren plateau (BP), which often emerges when a circuit's expressivity is so high that its ensemble of unitaries approximates a 2-design, leading to an exponentially decaying gradient~\cite{mcclean2018barren}.
The expressivity can be rigorously quantified by the dynamical Lie algebra (DLA) \cite{Ragone_2024, Fontana_2024}, $\mathfrak{L}$, which is the algebra generated by the circuit's Hamiltonians~\cite{d2007introduction}.
A crucial insight from quantum control theory is that if the dimension of the DLA, $\dim\mathfrak{L}$, scales polynomially with the number of qubits $n$, the circuit is protected from such expressivity-induced global BPs~\cite{larocca2021diagnosing}.

For our architecture in Figure~\ref{fig:vqc_graph}, all generators are strictly local, acting on at most two adjacent qubits.
Consequently, their nested commutators can only generate operators whose support grows contiguously. This algebraic bottleneck is the primary reason our PQC architecture is inherently trainable. As illustrated in Extended Data \ref{fig:dla_scaling}, the DLA dimension of our linear-chain PQC scales polynomially with the number of qubits $n$, in contrast to the exponential growth seen in globally-entangled circuits that leads to barren plateaus.

\subsubsection*{Causal structure and optimal circuit depth}

This algebraic constraint has a direct physical interpretation: the propagation of information is limited by a causal light cone.
Governed by the Lieb–Robinson bound, information in a system with local interactions propagates at a finite velocity~\cite{lieb1972finite}.

In our PQC, each block of entanglers expands the support of a local operator by $E$ sites in each direction. After $L$ blocks, an operator initially on a single qubit has an effective support of at most $2EL+1$ qubits (Figure~\ref{fig:lightcone_n16_E1}).

For the PQC to model system-wide correlations, this light cone must span the entire $n$-qubit chain. 
This condition of full causal connectivity is met when $2EL + 1 \ge n$, which defines a minimal depth for sufficient expressivity:
\begin{equation}
    L_{\mathrm{opt}} = \left\lceil \frac{n-1}{2E} \right\rceil.
    \label{eq:Lopt}
\end{equation}

For our design with $n=16$ and $E=1$, this yields an optimal depth of $L_{\mathrm{opt}}=8$, as depicted in Figure~\ref{fig:lightcone_n16_E1}.

\subsubsection*{Discussion: the trainability-expressivity sweet spot}

Equation~\eqref{eq:Lopt} provides a principled guideline for the design of PQCs by defining an optimal circuit depth that balances expressivity against trainability. 
Circuits with a depth $L < L_{\mathrm{opt}}$ are causally disconnected, preventing them from modeling long-range correlations for complex problems. 
In contrast, for depths $L \gg L_{\mathrm{opt}}$, the theoretical expressivity of the circuit (DLA dimension) saturates, and additional layers only introduce parameter redundancy and increased sensitivity to noise, which complicates optimization. 
This trade-off establishes an optimal operating point for maximum resource efficiency. Further details of the proof are provided in the Supplementary Information.

Based on this theoretical framework, we established the PQC architecture for the quantum sub-generators. Our design, with $n=16$ and $L=8$, is therefore positioned at this sweet spot: the minimum depth required for system-wide expressivity while retaining the algebraic structure that guarantees trainability.

\section*{Data availability}
All datasets analysed in this study are publicly available. The ISIC 2019 dermoscopic image dataset can be accessed from the ISIC Challenge archive (https://challenge.isic-archive.com/data/) under a CC BY-NC 4.0 license. The ODIR-5k ocular fundus image dataset is available through Kaggle (https://www.kaggle.com/datasets/andrewmvd/ocular-disease-recognition-odir5k). The RetinaMNIST dataset is part of the MedMNIST collection and can be downloaded at https://medmnist.com/, released under a CC BY 4.0 license (except DermaMNIST, which is CC BY-NC 4.0). No new human participant data were collected in this study.

\section*{Code availability}
The code is open-sourced at https://github.com/QingyueJ-nd/MediQ-GAN.git.

\section*{Author information}
\subsection*{Authors and Affiliations}

\noindent\textbf{Department of Computer Science and Engineering, University of Notre Dame, Notre Dame, IN, USA}

\noindent Qingyue Jiao \& Yiyu Shi

\noindent\textbf{Department of Computer Science, Boise State University, Boise, ID, USA}

\noindent Jun Zhuang

\noindent\textbf{Computer Science Department, University of California, Los Angeles, CA, USA}

\noindent Jason Cong

\noindent\textbf{Independent Researcher}

\noindent Yongcan Tang

\subsection*{Author Contributions}
Q.J. designed and implemented the experiments, performed latent geometry analysis, and wrote the first draft with Y.T. Y.T. developed the theoretical framework for the trainable parameterized quantum circuit architecture and derived the theoretical proofs in the Supplementary Information. J.Z. guided the GAN architecture design, J.C. advised on latent geometry interpretation, and Y.S. provided direction on medical imaging applications. All authors discussed the results, contributed to manuscript revisions, and approved the final version. J.Z., J.C., and Y.S. jointly supervised the work.

\subsection*{Corresponding author}
Correspondence to Yiyu Shi.

\section*{Funding declaration}
This study received no funding.


\bibliography{sn-bibliography}

\clearpage

\section*{Extended Data}\label{extended_data}

\setcounter{figure}{0} 
\captionsetup[figure]{name={Extended Data Fig.}} 

\setcounter{table}{0} 
\captionsetup[table]{name={Extended Data Table.}} 

\begin{figure*}[!htbp]
\centering
\includegraphics[width=\linewidth]{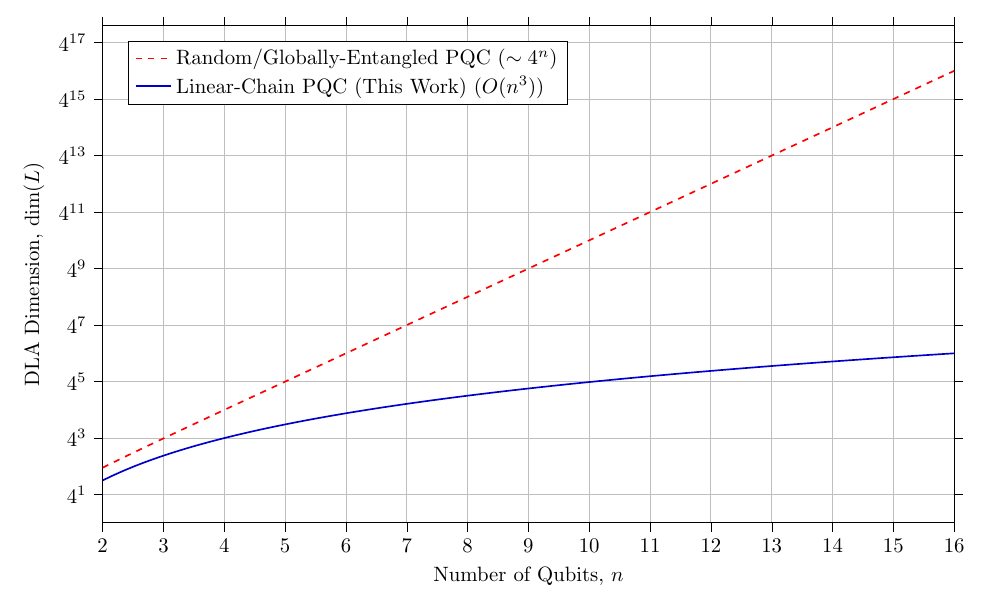} 
\caption{
    \textbf{PQC Trainability versus DLA Dimension Scaling.}
    This figure compares the scaling behavior of the DLA dimension for different PQC architectures.
    \textbf{(Dashed line) Globally-entangled PQC:} The dimension grows exponentially ($\sim 4^n$), leading to the barren plateau problem.
    \textbf{(Solid line) Linear-chain PQC (This work):} The dimension grows polynomially ($O(n^3)$), effectively avoiding barren plateaus and ensuring trainability.
}
\label{fig:dla_scaling} 
\end{figure*}

\begin{figure}[h!]
\centering
\begin{tikzpicture}

    \def\n{16} 
    \def\E{1}  
    \pgfmathsetmacro{\Lopt}{ceil((\n-1)/(2*\E))} 

    \begin{axis}[
        width=11cm,
        height=6cm,
        xlabel={Qubit Index, $i$},
        ylabel={Circuit Depth, $L$},
        xmin=-0.5, xmax={\n-0.5},
        ymin=0, ymax={\Lopt + 1},
        xtick={0, \n-1},
        xticklabels={$q_0$, $q_{\n-1}$},
        ytick={\Lopt},
        yticklabels={$L_{\mathrm{opt}}=\Lopt$},
        axis lines=left,
        clip=false,
    ]

    \addplot[only marks, mark=*, mark size=1.2pt] table {
        0  1  2  3  4  5  6  7  8  9  10  11  12  13  14  15
    };

    \fill[blue!15, opacity=0.8] 
        (axis cs: 0, 0) -- 
        (axis cs: 0, \Lopt) -- 
        (axis cs: {2*\E*\Lopt}, \Lopt) -- cycle;

    \draw[dashed, blue!65, very thick] 
        (axis cs: 0, \Lopt) -- (axis cs: \n-1, \Lopt)
        node[midway, above=3pt, black] {\small $2EL+1 \geq n$};

    \node[fill=white, draw, rounded corners, align=left, anchor=south east]
        at (axis description cs: 0.98, 0.02) {
        \(\displaystyle L_{\mathrm{opt}}=\left\lceil \frac{n-1}{2E}\right\rceil\)\\
        For \(n=\n,\,E=\E:\ L_{\mathrm{opt}}=\Lopt\)
    };

    \end{axis}
\end{tikzpicture}
\caption{
    \textbf{Light-cone growth and optimal depth in a 1D nearest-neighbor PQC.}
    This figure illustrates the causal light-cone expansion in a PQC with $n=16$ qubits and $E=1$. To achieve full causal connectivity, information from the first qubit ($q_0$) must be able to reach the last ($q_{15}$), requiring a minimum depth of $L_{\mathrm{opt}} = 8$.
}
\label{fig:lightcone_n16_E1}
\end{figure}

\begin{figure*}[!htbp]
  \centering
  \includegraphics[width=\textwidth]{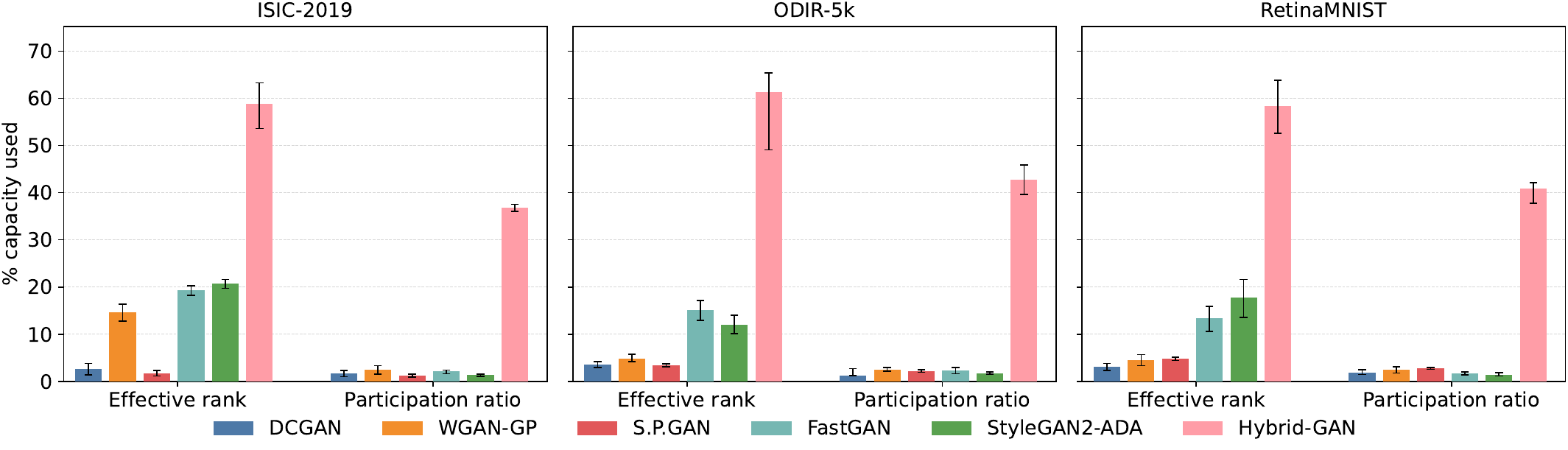}
  \caption{\textbf{Normalized effective rank (erank) and participation ratio (PR) for three datasets.}
  Values are normalized to the latent dimensionality (higher is better; greater fraction of latent directions used).
  Error bars indicate median with interquartile range (p25–p75).}
  \label{fig:erank_pr}
\end{figure*}

\begin{table}[!htbp]
\centering
\caption{\textbf{Size of benchmark datasets.}}
\label{tab:dataset-size}
\begin{tabular*}{\textwidth}{@{\extracolsep\fill}lcccccccc}
\toprule
\textbf{Dataset} & \multicolumn{8}{c}{\textbf{Classes}} \\
\midrule
\textbf{ISIC 2019} & mel & nv & bcc & akiec & bkl & df & vasc & scc \\
\midrule
train    & 822 & 2283 & 663 & 172 & 443 & 47 & 50 & 125 \\
test     & 205 & 571 & 166 & 43 & 111 & 12 & 13 & 31 \\
\midrule
\textbf{ODIR-5k} & N & D & G & C & A & H & M & O \\
\midrule
train    & 2298 & 1286 & 227 & 234 & 213 & 102 & 186 & 566 \\
test     & 575 & 322 & 57 & 59 & 53 & 26 & 46 & 142 \\
\midrule
\textbf{RetinaMNIST} & Grade 0 & Grade 1 & Grade 2 & Grade 3 & Grade 4 & -- & -- & -- \\
\midrule
train    & 571 & 149 & 261 & 226 & 74 &  &  &  \\
test     & 143 & 37 & 65 & 56 & 18 &  &  &  \\
\bottomrule
\end{tabular*}
\end{table}

\begin{table*}[!htbp]
\centering
\setlength{\tabcolsep}{2.5pt}
{\small
\caption{Minority vs. majority \textbf{macro-averaged} FID ($\downarrow$) by dataset and method.
Buckets use the median training class count within each dataset.}
\label{tab:fid-minor-major}
\begin{tabular*}{\textwidth}{@{\extracolsep\fill}l l c c}
\toprule
\textbf{Dataset} & \textbf{Method} & \textbf{Minority Avg} & \textbf{Majority Avg}  \\
\midrule
\multirow{8}{*}{\textbf{ISIC-2019}} 
  & DCGAN                           & 259.9 & 230.6  \\
  & WGAN-GP                         & 277.2 & 197.3  \\
  & Structure\mbox{-}preserving GAN & 283.7 & 245.3  \\
  & FastGAN                         & 230.8 & 213.4  \\
  & StyleGAN2\mbox{-}ADA            & 276.4 & 219.5  \\
  & DiT                             & 179.6 & \textbf{182.9} \\
  & Stable Diffusion 3              & 225.8 & 254.5  \\
  & \textbf{MediQ-GAN}                   & \textbf{174.5} & 189.0 \\
\midrule
\multirow{8}{*}{\textbf{ODIR-5k}} 
  & DCGAN                           & 217.9 & 226.3  \\
  & WGAN-GP                         & 200.4 & 203.4 \\
  & Structure\mbox{-}preserving GAN & 261.9 & 242.9  \\
  & FastGAN                         & 217.5 & 221.1  \\
  & StyleGAN2\mbox{-}ADA            & 235.2 & 228.9 \\
  & DiT                             & 181.0 & \textbf{152.6}  \\
  & Stable Diffusion 3              & 209.6 & 220.1  \\
  & \textbf{MediQ-GAN}                   & \textbf{158.9} & 165.9  \\
\midrule
\multirow{8}{*}{\textbf{RetinaMNIST}} 
  & DCGAN                           & 197.5 & 204.9  \\
  & WGAN-GP                         & 172.9 & 158.8 \\
  & Structure\mbox{-}preserving GAN & 295.8 & 220.7 \\
  & FastGAN                         & 186.9 & 191.6 \\
  & StyleGAN2\mbox{-}ADA            & 214.6 & 209.9  \\
  & DiT                             & \textbf{130.8} & 127.2  \\
  & Stable Diffusion 3              & 166.8 & 165.9  \\
  & MosaiQ                          & 376.12 & 331.86 \\
  & \textbf{MediQ-GAN}                   & 141.4 & \textbf{139.4} \\
\bottomrule
\end{tabular*}
} %
\end{table*}

\begin{table*}[!htbp]
\centering
\setlength{\tabcolsep}{2.5pt}
{\small
\caption{Minority vs. majority \textbf{macro-averaged intra-class LPIPS} ($\uparrow$) by dataset and method. Higher values indicate greater within-class perceptual diversity. }
\label{tab:lpips-minor-major}
\begin{tabular*}{\textwidth}{@{\extracolsep\fill}l l c c}
\toprule
\textbf{Dataset} & \textbf{Method} & \textbf{Minority Avg} & \textbf{Majority Avg} \\
\midrule
\multirow{9}{*}{\textbf{ISIC-2019}} 
  & Real (reference)                & 0.457 & 0.447 \\
  & DCGAN                           & 0.376 & 0.380 \\
  & WGAN\mbox{-}GP                  & 0.354 & 0.382 \\
  & Structure\mbox{-}preserving GAN & 0.333 & 0.346 \\
  & FastGAN                         & 0.422 & 0.414 \\
  & StyleGAN2\mbox{-}ADA            & 0.411 & 0.394 \\
  & DiT                             & 0.367 & 0.411 \\
  & Stable Diffusion 3              & 0.266 & 0.239 \\
  & \textbf{MediQ-GAN}                   & \textbf{0.409} & \textbf{0.417} \\
\midrule
\multirow{9}{*}{\textbf{ODIR-5k}} 
  & Real (reference)                & 0.612 & 0.594 \\
  & DCGAN                           & 0.583 & \textbf{0.597} \\
  & WGAN\mbox{-}GP                  & 0.590 & 0.584 \\
  & Structure\mbox{-}preserving GAN & 0.479 & 0.534 \\
  & FastGAN                         & \textbf{0.609} & 0.578 \\
  & StyleGAN2\mbox{-}ADA            & 0.578 & 0.550 \\
  & DiT                             & \textbf{0.609} & 0.560 \\
  & Stable Diffusion 3              & 0.522 & 0.525 \\
  & \textbf{MediQ-GAN}                   & 0.601 & 0.578 \\
\midrule
\multirow{9}{*}{\textbf{RetinaMNIST}} 
  & Real (reference)                & 0.396 & 0.372 \\
  & DCGAN                           & 0.217 & 0.195 \\
  & WGAN\mbox{-}GP                  & 0.344 & 0.353 \\
  & Structure\mbox{-}preserving GAN & 0.240 & 0.269 \\
  & FastGAN                         & \textbf{0.394} & 0.376 \\
  & StyleGAN2\mbox{-}ADA            & 0.424 & 0.350 \\
  & DiT                             & 0.321 & 0.322 \\
  & Stable Diffusion 3              & 0.336 & 0.333 \\
  & MosaiQ                          & 0.079 & 0.080 \\
  & \textbf{MediQ-GAN}                   & 0.384 & \textbf{0.378} \\
\bottomrule
\end{tabular*}
} 
\end{table*}

\setlength{\tabcolsep}{3pt} 
\begin{table*}[!htbp]
\caption{\textbf{Per-class FID ($\downarrow$) scores} of representative classes of ISIC 2019, ODIR-5k, and RetinaMNIST on real quantum hardware.}
\label{tab:fid-real}
\centering
\resizebox{\textwidth}{!}{%
\begin{tabular}{@{}l
   rrr 
   rrr 
   rrr 
   @{}}
\toprule
& \multicolumn{3}{c}{\textbf{ISIC 2019}} 
& \multicolumn{3}{c}{\textbf{ODIR-5k}} 
& \multicolumn{3}{c}{\textbf{RetinaMNIST}} \\
\cmidrule(lr){2-4}\cmidrule(lr){5-7}\cmidrule(lr){8-10}
\textbf{Method} 
& \textbf{mel} & \textbf{df} & \textbf{vasc}
& \textbf{N}   & \textbf{H}  & \textbf{M}
& \textbf{GR 0} & \textbf{GR 3} & \textbf{GR 4} \\
\midrule
Simulation
           & 187.3 & 182.2 & 171.4
           & 162.8 & 167.4 & 144.3
           & 127.3 & 159.7 & 116.9 \\
IBM Marrakesh
           & 256.3 & 261.5 & 214.3
           & 200.9 & 197.5 & 168.2
           & 146.3 & 197.6 & 177.7 \\
\bottomrule
\end{tabular}%
}
\end{table*}

\setlength{\tabcolsep}{4pt}
\renewcommand{\arraystretch}{1.12}
\begin{table*}[!htbp]
\small
\caption{\textbf{Log Condition Number for three datasets.}
Reported as median [p25–p75] across samples. 
$\log_{10}\kappa(J)$ measures generator conditioning (lower is better). }
\label{tab:erank_multi}
\centering
\begin{tabular*}{\textwidth}{@{\extracolsep{\fill}}lccc@{}}
\toprule
\textbf{Model} & $\textbf{ISIC 2019}$ & \textbf{ODIR-5k} & \textbf{RetinaMNIST} \\
\midrule
DCGAN          & $2.54\,[2.37,2.93]$ & $2.54\,[2.47,2.62]$ & $2.44\,[2.33,2.54]$  \\
WGAN-GP        & $4.39\,[4.27,4.56]$ & $4.61\,[4.56,4.66]$ & $4.81\,[4.72,4.88]$ \\
S.P.\,GAN      & $2.62\,[2.47,2.89]$ & $2.61\,[2.57,2.67]$ & $2.83\,[1.79,2.86]$ \\
FastGAN        & $4.85\,[4.71,5.04]$ & $3.39\,[3.14,3.43]$ & $3.83\,[3.70,4.18]$\\
StyleGAN2-ADA  & $5.78\,[5.53,6.24]$ & $6.25\,[5.59,6.90]$ & $6.58\,[5.71,6.88]$ \\
\textbf{MediQ-GAN} 
               & $\mathbf{2.09\,[1.44,2.47]}$ 
               & $\mathbf{2.04\,[1.26,2.25]}$ 
               & $\mathbf{2.01\,[1.79,2.38]}$  \\
\bottomrule
\end{tabular*}
\end{table*}

\end{document}